\documentclass[lettersize, journal]{IEEEtran}

\usepackage{amsmath,amsfonts,amssymb} 
\usepackage{graphicx} 
\usepackage{cite} 
\usepackage{algorithm}
\usepackage{algpseudocode}

\usepackage{array} 
\usepackage{booktabs} 
\usepackage{multirow} 
\usepackage{subcaption} 

\usepackage[table]{xcolor} 
\usepackage[normalem]{ulem}

\usepackage{filecontents}
\usepackage{flushend}

\definecolor{seagreen}{rgb}{0.18, 0.55, 0.34}
\definecolor{royalpurple}{rgb}{0.47,0.32,0.66}
\definecolor{brown(traditional)}{rgb}{0.59, 0.29, 0.0}
\definecolor{blue}{rgb}{0.3, 0.2, 0.9}
\usepackage[colorlinks,
            linkcolor=blue,
            anchorcolor=blue,
            citecolor=blue]{hyperref}

\DeclareMathOperator*{\argmax}{argmax}

\begin{document}

\title{MedAlign: A Synergistic Framework of Multimodal Preference Optimization and Federated Meta-Cognitive Reasoning}

\author{
Siyong Chen*, Jinbo Wen*, Jiawen Kang, Tenghui Huang, Xumin Huang, Yuanjia Su, Hudan Pan, Zishao Zhong,\\ Dusit Niyato, \textit{Fellow, IEEE}, Shengli Xie, \textit{Fellow, IEEE}, and Dong In Kim, \textit{Life Fellow, IEEE}

\thanks{
S. Chen, J. Kang, T. Huang, X. Huang, Y. Su, and S. Xie are with the School of Automation, Guangdong University of Technology, Guangzhou, China (e-mails: 3122000875@mail2.gdut.edu.cn, kavinkang@
gdut.edu.cn, 3123000938@mail2.gdut.edu.cn, huangxumin@gdut.edu.cn, syj1216902331@163.com, shlxie@gdut.edu.cn).
J. Wen is with the College of Computer Science and Technology, Nanjing University of Aeronautics and Astronautics, Nanjing, China (e-mail: jinbo1608@nuaa.edu.cn). 
H. Pan and Z. Zhong are with State Key Laboratory of Traditional Chinese Medicine Syndrome, The Second Affiliated Hospital of Guangzhou University of Chinese Medicine, Guangdong Provincial Hospital of Chinese Medicine, Guangdong Provincial Academy of Chinese Medical Sciences, Guangzhou, China, and Chinese Medicine Guangdong Laboratory, Zhuhai, China (e-mails: hdpan@gzucm.edu.cn, zhongzishao@gzucm.edu.cn).
D. Niyato is with the College of Computing and Data Science, Nanyang Technological University, Singapore (e-mail: dniyato@ntu.edu.sg).
D. I. Kim is with the Department of Electrical and Computer Engineering, Sungkyunkwan University, South Korea (e-mail: dikim@skku.ac.kr).
* means equal contribution. (\textit{Corresponding authors: Jiawen Kang and Shengli Xie})

}

}

\maketitle

\begin{abstract}
Recently, large models have shown significant potential for smart healthcare. However, the deployment of Large Vision-Language Models (LVLMs) for clinical services is currently hindered by three critical challenges: a tendency to hallucinate answers not grounded in visual evidence, the inefficiency of fixed-depth reasoning, and the difficulty of multi-institutional collaboration. To address these challenges, in this paper, we develop MedAlign, a novel framework to ensure visually accurate LVLM responses for Medical Visual Question Answering (Med-VQA). Specifically, we first propose a multimodal Direct Preference Optimization (mDPO) objective to explicitly align preference learning with visual context. We then design a Retrieval-Aware Mixture-of-Experts (RA-MoE) architecture that utilizes image and text similarity to route queries to a specialized and context-augmented LVLM (i.e., an expert), thereby mitigating hallucinations in LVLMs. To achieve adaptive reasoning and facilitate multi-institutional collaboration, we propose a federated governance mechanism, where the selected expert, fine-tuned on clinical datasets based on mDPO, locally performs iterative Chain-of-Thought (CoT) reasoning via the local meta-cognitive uncertainty estimator. Extensive experiments on three representative Med-VQA datasets demonstrate that MedAlign achieves state-of-the-art performance, outperforming strong retrieval-augmented baselines by up to $11.85\%$ in F1-score, and simultaneously reducing the average reasoning length by $51.60\%$ compared with fixed-depth CoT approaches.
\end{abstract}


\section{Introduction}
\label{sec:introduction}

Large Vision-Language Models (LVLMs) have emerged as a promising frontier, holding transformative potential for augmenting clinical services and streamlining diagnostic workflows in smart healthcare~\cite{11153787}. Their ability to synthesize multimodal data offers an unprecedented opportunity to provide more accurate, efficient, and personalized medical diagnosis services. However, the transition of these powerful models from research benchmarks to reliable deployment in real medical environments remains a challenging process. This gap between potential and practical deployment arises from fundamental limitations in architectural design, reasoning capabilities, and data acquisition paradigms. Specifically, the widespread adoption and trustworthiness of LVLMs are currently hampered by three significant and interconnected challenges: (I) \textit{Lack of visual grounding}, which leads to factual hallucinations in generated responses contradict the underlying visual evidence~\cite{Favero_2024_CVPR, ye2025claim}; (II) \textit{Inefficiency of fixed-depth reasoning}, which squanders computational resources on simple queries while failing to allocate sufficient reasoning depth for complex queries~\cite{chen2024, khademsohi2024selfxit}; (III) \textit{Inability to facilitate multi-institutional collaboration}. This limitation hinders the development of LVLMs that transcend the constraints of single-institution datasets, particularly in medical scenarios~\cite{10976244}. The above challenges collectively lead to unreliable reasoning and inefficiency of LVLMs in smart healthcare.

For these challenges, existing studies merely offer partial solutions. For \textit{Challenge I}, mainstream alignment techniques such as Direct Preference Optimization (DPO) are primarily text-centric, often causing LVLMs to favor linguistic plausibility over visual fidelity. Even reasoning methods with visual perception capabilities struggle to fully bridge this gap~\cite{zhang-etal-2025-improve}. Although Retrieval-Augmented Generation (RAG) can provide external knowledge, it often acts as a passive addition rather than an integrated component~\cite{10812735}. For \textit{Challenges II} and \textit{III}, adaptive reasoning techniques have been explored to mitigate the inefficiency of static computational graphs~\cite{yue2025dots, yuan2024instanceadaptive}. Nevertheless, the current work lacks a robust mechanism to regulate this dynamic process. Although federated learning has become the established standard for distributed training~\cite{carrillo2024fedcbo}, extending its principles to enable collaborative reasoning among LVLMs remains a key unresolved challenge~\cite{zheng2025fedcalm}.

To this end, in this paper, we propose MedAlign, a unified framework to ensure reliable LVLM responses for Medical Visual Question Answering (Med-VQA). Specifically, to address Challenge I, we propose a multimodal DPO (mDPO) objective to directly penalize visual incongruity and a Retrieval-Aware Mixture-of-Experts (RA-MoE) architecture that transforms retrieval from a passive attachment into active reasoning guidance. Furthermore, to address Challenges II and III, we propose a federated governance mechanism that employs meta-cognitive uncertainty estimators to autonomously control the depth of Chain-of-Thought (CoT) reasoning, thereby enhancing diagnosis efficiency. Moreover, it extends federation principles from training to inference, establishing a paradigm for multi-institutional collaborative reasoning that current paradigms lack. The main contributions of this paper are summarized as follows:
\begin{itemize}
    \item \textbf{Multimodal Direct Preference Optimization Objective.} Unlike traditional DPO objectives, which often favor linguistic plausibility over visual fidelity, we propose a novel mDPO objective to directly mitigate hallucinations in LVLMs for Med-VQA. This objective explicitly penalizes visual incongruity through a cross-modal preference loss, and the training process is reinforced by a data-driven anchor to ensure robust and stable model performance. (For Challenge I)
    \item \textbf{Retrieval-Aware Mixture-of-Experts Architecture.} Unlike conventional approaches that either treat retrieved evidence as passive input or rely solely on embedding-based routing, we propose an RA-MoE architecture with zero-shot routing, which directly translates multimodal retrieval scores into gating signals. The proposed architecture transforms retrieval from passive input attachments into active reasoning guidance, dynamically activating a specialized and context-augmented LVLM (i.e., an expert) that best aligns with the joint image–text semantics of the query. (For Challenge I)
    \item \textbf{Federated Governance for Adaptive Reasoning.} We innovatively propose a federated governance mechanism for adaptive reasoning, extending the principles of federation beyond distributed training to inference. By employing local meta-cognitive uncertainty estimators, the selected expert, fine-tuned on clinical datasets via mDPO, can autonomously regulate its CoT reasoning depth, thereby providing a controllable and efficient alternative to conventional prompting. (For Challenges II and III)
    \item \textbf{State-of-the-Art Performance of MedAlign.} We conduct a comprehensive evaluation on three representative Med-VQA benchmarks. Simulation results demonstrate that MedAlign achieves new State-of-the-Art (SOTA) performance, outperforming retrieval-augmented baselines by up to $11.85\%$ in F1-score on average. Moreover, MedAlign exhibits superior robustness to noise and reduces computational overhead by $51.60\%$ through the proposed federated governance mechanism. 
\end{itemize}

The rest of this paper is organized as follows: 
Section~\ref{sec:related_work} reviews related studies on RAG, MoE routing, and adaptive reasoning. 
Section~\ref{sec:methodology} presents the framework design of MedAlign, combining a visual-faithful mDPO objective, an RA-MoE architecture, and the federated governance mechanism for adaptive reasoning. Section~\ref{sec:experiments} presents comprehensive experiments on Med-VQA benchmarks to evaluate the performance, robustness, and efficiency of MedAlign. 
Finally, Section~\ref{sec:conclusion} concludes the paper. The main mathematical notations of this paper are illustrated in Table~\ref{Notation}.

\begin{table}[t]
\renewcommand{\arraystretch}{1.6}
\caption{Key mathematical notations of this paper.}
\begin{tabular}{c|m{6.6cm}} 
\toprule[1.3pt]
\rowcolor{gray!10}
\multicolumn{1}{c|}{\textbf{Notations}}  & \multicolumn{1}{c}{\textbf{Definition}} \\ \hline
\(\pi_{\theta}\) & Policy model parameterized by \(\theta\) \\ \hline
\(\pi_{\mathrm{ref}}\) & Frozen reference model used for DPO \\ \hline
\(E_{d}\) & Expert specialized for domain \(d\) \\ \hline
\(E_{\mathrm{retrieval}}\) & Frozen multimodal encoder used for retrieval and routing \\ \hline
\(\mathcal{D}_{\mathrm{pref}}\) & Preference dataset of tuples \((I, Q, A^{+}, A^{-})\) \\ \hline
\(\mathcal{D}_{\mathrm{CM}}\) & Cross-modal preference dataset (visual/text pairings and preference labels) \\ \hline
\(\hat{r}_{\theta}(x,y)\) & Implicit reward for response \(y\) given input \(x\) \\ \hline
\(\mathcal{L}_{\mathrm{DPO}}\) & DPO loss \\ \hline
\(\mathcal{L}_{\mathrm{CM}}\) & Cross-modal preference loss enforcing visual grounding \\ \hline
\(q_{\mathrm{multi}}\) & Multimodal query embedding \\ \hline
\(s_{d}\) & Raw aggregated retrieval score associated with domain \(d\) \\ \hline
\(P(d|I,Q)\) & Probability of routing the query \((I,Q)\) to expert \(d\) \\ \hline
\(h_{t}\) & Decoder hidden state at CoT step \(t\) \\ \hline
\(\Delta_{\mathrm{stability}}\) & Stability adjustment term \\ \hline
\(c_{i,t}\) & Stability-adjusted confidence score for instance \(i\) at step \(t\) \\ \hline
\(\gamma\) & Confidence threshold used in the halting decision (instance is halted if \(c_{i,t}\geq\gamma\)) \\
\bottomrule[1.3pt]
\end{tabular}\label{Notation}
\end{table}

\section{Related Work}
\label{sec:related_work}

\subsection{Visually-Grounded Preference Optimization}
The field of Med-VQA has transitioned from discriminative models to Large Language Models (LLMs)~\cite{11153787, 11016939}. While these advanced LLMs perform well on open-ended queries across diverse medical scenarios, a persistent challenge in practical clinical applications is their tendency toward factual hallucination. To address this issue, researchers have widely adopted alignment techniques, such as Reinforcement Learning from Human Feedback (RLHF)~\cite{dai2024safe} and DPO~\cite{zeng2024}, to fine-tune LLMs. For instance, Dai et al.~\cite{dai2024safe} proposed an algorithm called safe RLHF, which concurrently optimizes helpfulness and safety using multimodal rewards and constrained optimization for multimodal alignment. In the realm of DPO, Zeng et al.~\cite{zeng2024} proposed token-level DPO to align LLMs with human preferences by optimizing policy at the token level. However, a significant limitation arises when these text-centric preference optimization techniques are directly applied to multimodal tasks like Med-VQA. In such scenarios, the optimization process tends to prioritize linguistic fluency and plausibility over strict visual accuracy, potentially leading to visually ungrounded or hallucinatory outputs. Recognizing this critical gap, recent efforts have explored fine-grained adaptation mechanisms to align multimodal reasoning with visual evidence. For example, Wang et al.~\cite{wang-etal-2025-aspo} proposed an adaptive sentence-level preference optimization framework that enhances multimodal reasoning by dynamically modeling fine-grained textual-visual preferences. Li et al.~\cite{li-etal-2025-generate} introduced a self-evolutionary approach to improve context faithfulness through iterative fine-grained sentence-level optimization and discriminative verification. 

Within the clinical services domain, specialized preference optimization strategies have been developed to align LVLMs with domain-specific diagnostic criteria and expert knowledge. For instance, Zhu et al.~\cite{DBLP:journals/corr/abs-2412-06141} proposed a multimodal clinical preference optimization approach, studying how to incorporate diagnostic and clinical rules into preference signals, thereby enhancing the alignment of LVLMs. While these efforts substantially enhance the robustness of LVLMs, they often address visual grounding only indirectly or emphasize domain-specific rule compliance instead of establishing grounding in fundamental visual evidence.

\subsection{Knowledge-Augmented Reasoning}
Traditional RAG methods often treat retrieved knowledge as passive input. To overcome this limitation, emerging multimodal RAG systems pursue tighter integration between retrieval and reasoning. For instance, Yu et al.~\cite{yu2025visrag} used VLMs to directly retrieve documents as visual objects, thereby avoiding information loss in text parsing and tightly coupling retrieval with visual features. Xiao et al.~\cite{xiao2024mmedrag} proposed a multimodal RAG system for medical VLMs. This system integrates domain-aware retrieval with adaptive context selection and fine-grained multimodal reasoning, improving factual accuracy across multiple clinical datasets. Beyond traditional dense models, MoE architectures improve task specialization and computational efficiency by activating only a subset of experts for each input. Building on this idea, integrating retrieval mechanisms with MoE has emerged as a promising research direction. For example, Lee et al.~\cite{Lee2024RouterRetrieverRO} developed a mixture-of-task-experts framework for embedding models, where task-specific expert parameters are dynamically activated to improve representation quality. Li et al.~\cite{li2025ragddr} made the retrieval process differentiable and integrated retrieval quality into the loss function, thereby aligning retrieval more tightly with downstream generation objectives. These approaches align well with the idea of integrating retrieval scores to drive MoE routing. However, existing multimodal RAG and MoE approaches either treat retrieved evidence as passive input or rely solely on embedding-based routing, leaving open critical gaps in context-aware expert selection, robustness to retrieval noise, and efficient adaptive inference.


\subsection{LLM Inference under Meta-Cognitive Uncertainty}
The efficiency of reasoning processes in LLMs remains a critical challenge, particularly in medical scenarios. To mitigate the computational burden in LLMs caused by static reasoning depths of CoT prompting, recent studies have proposed improved CoT prompting techniques. For instance, Yuan et al.~\cite{yuan2024instanceadaptive} proposed an instance-adaptive zero-shot CoT strategy that dynamically determines when to generate further reasoning steps and when to terminate, providing a practical solution to optimize computational resources. Similarly, Yue et al.~\cite{yue2025dots} proposed an approach that searches for optimal reasoning trajectories, thereby enabling LLMs to dynamically adjust reasoning depth and steps. While these approaches effectively alleviate computational inefficiency, determining the optimal termination point remains a challenging problem. Related research has explored self-truncation mechanisms in LLMs to mitigate excessively long reasoning chains~\cite{yue2025dots}, as well as scalable system-level implementations of early-exit strategies for improving reasoning efficiency~\cite{chen2024}.
To effectively manage adaptive reasoning, a robust method for quantifying model uncertainty is essential. Current research has distinguished between aleatoric uncertainty (i.e., inherent data noise) and epistemic uncertainty (i.e., knowledge insufficiency within LLMs), where a high level of epistemic uncertainty typically indicates the need for deeper or extended reasoning~\cite{wimmer2023quantifying}.


\section{Framework Design of MedAlign}
\label{sec:methodology}

The architecture of MedAlign consists of three technical components: the mDPO objective, the RA-MoE architecture with zero-shot routing, and the federated governance mechanism for adaptive CoT reasoning. We detail each of these components as follows:

\subsection{Multimodal Direct Preference Optimization}
\label{subsec:mdpo}


A central challenge in clinical VQA is ensuring that generated answers are not only linguistically plausible but also factually and visually grounded in the provided medical images. Although standard alignment techniques such as DPO are effective for general alignment tasks, they fail to explicitly account for modality-specific constraints or visual grounding requirements.
To address the challenges of visual grounding and training stability, we propose mDPO, a new learning objective that enhances DPO along two orthogonal dimensions. 

Without loss of generality, we first define the preference dataset as
\(\mathcal{D}_{\mathrm{pref}}=\{(x_i,y_{w,i},y_{l,i})\}_{i=1}^N\), where $N$ represents the total number of tuples in the preference dataset. Each multimodal input \(x=(I,Q)\) consists of an image \(I\) and a corresponding question \(Q\). Here, \(y_w\) and \(y_l\) denote the preferred answer and the less-preferred answer, respectively. The standard DPO loss aligns a policy model \(\pi_\theta\) with human preferences by maximizing the margin between the implicit reward of a preferred answer \(y_w\) and an unsatisfied answer \(y_l\), which is given by
\begin{equation}
\label{eq:dpo_loss}
\begin{split}
\mathcal{L}_{\mathrm{DPO}} ={}& -\mathbb{E}_{(x, y_w, y_l) \sim \mathcal{D}_{\mathrm{pref}}} \left[ \log \sigma \left( \hat{r}_\theta(x, y_w) \right. \right. \\
                             & \left. \left. - \hat{r}_\theta(x, y_l) \right) \right],
\end{split}
\end{equation}
where \(\hat{r}_\theta(x, y) = \beta \log(\pi_\theta(y|x)/\pi_{\mathrm{ref}}(y|x))\), and $\pi_{\mathrm{ref}}$ denotes a fixed reference distribution. Here, the scalar $\beta > 0$ serves as an inverse temperature parameter, and a larger $\beta$ indicates the small probability differences between responses, while a smaller $\beta$ compresses the reward scale. For clinical multimodal VQA, we employ a frozen multimodal pre-trained LVLM. In clinical contexts, domain-aware preference design and clinical relevance filtering have been shown to substantially improve alignment and reduce clinical hallucinations~\cite{DBLP:journals/corr/abs-2412-06141}.

Eq. (\ref{eq:dpo_loss}) remains blind to the visual modality, as it does not explicitly incorporate visual evidence into the alignment process. If \(\pi_\theta(y|x)\) and \(\pi_{\mathrm{ref}}(y|x)\) do not depend on the image component of \(x=(I,Q)\), then for two different images \(I_1\neq I_2\), we have
\(
\hat r_\theta((I_1,Q),A)=\hat r_\theta((I_2,Q),A),
\)
indicating that the reward signal is invariant to image changes and provides no gradient to penalize visually incongruent answers.

\subsubsection{Cross-modal preference for visual grounding}
To produce a direct and testable training signal that forces the policy to use visual evidence, we adopt a cross-modal preference loss~\cite{amini-etal-2024-direct}. Specifically, for a given QA pair $(Q,A)$, we construct an image pair $(I_w,I_l)$, where $I_w$ is the supporting image that provides visual evidence consistent with $A$, and $I_l$ is the contradictory image that lacks or conflicts with such visual support. The cross-modal preference loss enforces a reward margin between the two image contexts, as given by
\begin{equation}
\mathcal{L}_{\mathrm{CM}}\ = -\mathbb{E} \left[ \log \sigma \left( \hat{r}_\theta(I_w, Q, A) - \hat{r}_\theta(I_l, Q, A) \right) \right],
\label{eq:cm_loss}
\end{equation}
where $\hat r_\theta(I,Q,A)=\beta\log\frac{\pi_\theta(A| I,Q)}{\pi_{\mathrm{ref}}(A| I,Q)}$. This method constructs image–contrast pairs for the same text and enforces a reward margin between supporting and contradicting images, enhancing visual grounding and reducing hallucinations~\cite{wang2024mdpoconditionalpreferenceoptimization, Favero_2024_CVPR}.

\begin{algorithm}[t]
\small
\caption{mDPO Training}
\label{alg:mdpo_training}
\begin{algorithmic}[1] 

\State \textbf{Input:} Policy model \(\pi_\theta\), reference model \(\pi_{\mathrm{ref}}\), datasets \(\mathcal{D}_{\mathrm{pref}}, \mathcal{D}_{\mathrm{CM}}\), and hyperparameters \(\beta, \lambda_{\mathrm{CM}}, \lambda_{\mathrm{RA}}, \eta\).
\State \textbf{Output:} Optimized parameters \(\theta^*\).
\Statex
Compute the anchor \(\delta\) according to Eq.~\eqref{eq:-model reward distribution}.

\Statex
Initialize an optimizer for \(\theta\).

\For{each training epoch}
    \For{each mini-batch \(B_{\mathrm{pref}}, B_{\mathrm{CM}}\) from the datasets}
        
        \State Compute batch loss \(\mathcal{L}_{\mathrm{DPO}}\) on \(B_{\mathrm{pref}}\) using Eq.~\eqref{eq:dpo_loss}.
        \State Compute batch loss \(\mathcal{L}_{\mathrm{RA}}\) on \(B_{\mathrm{pref}}\) using Eq.~\eqref{eq:ra_loss}.
        \State Compute batch loss \(\mathcal{L}_{\mathrm{CM}}\) on \(B_{\mathrm{CM}}\) using Eq.~\eqref{eq:cm_loss}.
        
        \State \textit{\textcolor{blue}{\#\#\# Weighted loss aggregation \#\#\#}}
        \State \(\mathcal{L}_{\mathrm{total}} \gets \mathcal{L}_{\mathrm{DPO}} + \lambda_{\mathrm{RA}}\mathcal{L}_{\mathrm{RA}} + \lambda_{\mathrm{CM}}\mathcal{L}_{\mathrm{CM}}\).

        \State Update parameters \(\theta\) by descending the gradient of \(\mathcal{L}_{\mathrm{total}}\).
        
    \EndFor
\EndFor
\State \Return \(\theta\)
\end{algorithmic}
\end{algorithm}

\subsubsection{Anchor-based reward regularization for stability}
\label{subsubsec:arr}

To address training instability, we introduce an \emph{Anchor-based Reward Regularization} (ARR) term that enforces an absolute reward baseline for preferred examples. ARR augments the standard DPO margin with a data-driven anchor \(\delta\), yielding a stabilized preference objective. A direct way to incorporate the anchor is to require that the implicit reward margin exceed \(\delta\). Hence, the reward-anchored DPO loss is expressed as
\begin{equation}
\label{eq:ra_loss}
\begin{split}
\mathcal{L}_{\mathrm{RA}} ={}& -\mathbb{E}_{(x, y_w, y_l) \sim \mathcal{D}_{\mathrm{pref}}} \left[ \log \sigma \left( \hat{r}_\theta(x, y_w) \right. \right. \\
                             & \left. \left. - \hat{r}_\theta(x, y_l) - \delta \right) \right].
\end{split}
\end{equation}
The anchor term is inspired by recent DPO calibration studies~\cite{wang2024mdpoconditionalpreferenceoptimization, amini-etal-2024-direct}, which aim to address issues of reward scaling and training instability. To ensure that \(\delta\) is unbiased, we estimate it using a separate calibration set \(\mathcal{D}_{\mathrm{pref}}^{\mathrm{cal}}\), which consists of a subset of preference pairs held out from training updates to prevent evaluation-training leakage. The anchor is defined as an empirical percentile of the reference-model reward distribution, which is given by
\begin{equation}
\label{eq:-model reward distribution}
\delta = \mathrm{Prc}_q\Big(\big\{\hat r_{\mathrm{ref}}(x,y_w)\mid(x,y_w,y_l)\in\mathcal{D}_{\mathrm{pref}}^{\mathrm{cal}}\big\}\Big),
\end{equation}
where \(\mathrm{Prc}_q\) denotes the empirical \(q\)-th percentile.

\begin{figure*}[t]
    \centering
    \includegraphics[width=0.92\textwidth]{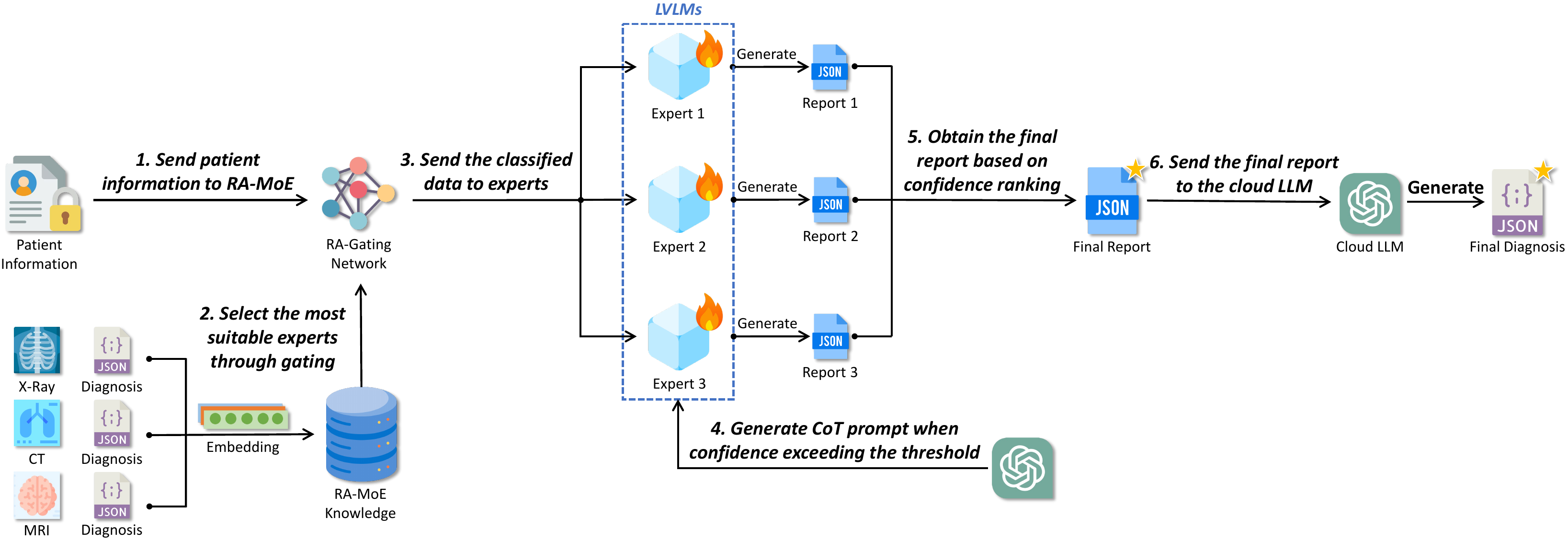}
    \caption{The overview of the LVLM inference paradigm in MedAlign, which operates in two phases. \textit{Phase 1: Retrieval-aware expert routing.} The input query $(I, Q)$ is encoded into a multimodal embedding, which is used to perform parallel retrieval across $D$ domain knowledge bases. The resulting relevance scores serve as a zero-shot gating signal to select the optimal expert $E_{d*}$, while the retrieved textual context from the same operation provides additional knowledge augmentation. \textit{Phase 2: federated governance for adaptive reasoning.} The selected expert then performs iterative CoT reasoning in parallel across $N$ federated sites. At each step, a local meta-cognitive uncertainty estimator $g_{\psi}$ determines whether to continue reasoning or to report its confident result to a central server. The server aggregates these high-confidence reasoning chains and synthesizes the final answer $A^*$ once a quorum consensus is achieved.}
    \label{fig:framework_overview}
\end{figure*}

\subsubsection{Multimodal preference optimization objective}

The proposed mDPO objective is a composite loss that holistically trains LVLMs, which is a linearly weighted sum of the three component losses:
\begin{equation}
\mathcal{L}_{\mathrm{mDPO}} = \mathcal{L}_{\mathrm{DPO}} + \lambda_{\mathrm{CM}} \mathcal{L}_{\mathrm{CM}} + \lambda_{\mathrm{RA}} \mathcal{L}_{\mathrm{RA}},
\label{eq:mdpo_final}
\end{equation}
where \(\lambda_{\mathrm{CM}}\) and \(\lambda_{\mathrm{RA}}\) are balancing hyperparameters. Algorithm \ref{alg:mdpo_training} outlines the overall training pipeline of LVLMs using the proposed mDPO objective. Its computational complexity is \(\mathcal{O}(B_{\mathrm{pref}} T + B_{\mathrm{CM}} T_{\mathrm{image}})\), where \(B_{\mathrm{pref}}\) and \(B_{\mathrm{CM}}\) are the batch sizes for the preference and cross-modal datasets, respectively, \(T\) is the time steps for processing text input, and \(T_{\mathrm{image}}\) is the time steps for processing image input.

Once trained via mDPO, LVLMs are deployed within a sophisticated inference paradigm designed to coordinate knowledge-intensive reasoning, as illustrated in Fig.~\ref{fig:framework_overview}. This paradigm consists of two key components: an RA-MoE architecture that selects and augments the optimal domain expert, and a federated governance mechanism that manages the adaptive reasoning process of the selected expert, employing a novel meta-cognitive inference module to ensure that the final results are robust and well-grounded.

\subsection{Retrieval-Aware Mixture-of-Experts}
\label{ssubsec:ra_moe}
MoE is a machine learning technique consisting of a gating network and multiple expert networks~\cite{chen2022towards}. Each expert specializes in processing a particular subset of the input data, while the gating network selectively activates only a subset of experts for each input. Although trainable gating networks can achieve strong performance by learning optimal expert assignments, they require specialized supervision signals for routing and introduce extra parameters~\cite{chen2022towards}, increasing model complexity and reducing interpretability in the decision-making process. To enhance interpretability during inference, we propose a novel unsupervised expert routing architecture, which leverages pre-trained multimodal retrieval models to directly transform retrieval relevance scores into expert gating signals~\cite{Lee2024RouterRetrieverRO, zhou-etal-2024-unveiling}, thereby eliminating the need for routing-label supervision and making the routing process fully transparent and interpretable.

\subsubsection{Domain knowledge base construction}
We construct domain-specific vector databases, denoted as \(\{\mathcal{DB}_d\}_{d=1}^D\). Considering that defining clear knowledge boundaries in the medical domain is a non-trivial challenge~\cite{bhattarai-etal-2024-document}, we adopt the unified medical language system for standardized domain delineation~\cite{diagnostics14111204}. Specifically, we map each document in our corpus to one or more medical concepts. Based on the association of these concepts with our \(D\) predefined clinical specialties, the document index is then added to one or more corresponding databases \(\mathcal{DB}_d\). This strategy accounts for the inherently cross-domain nature of medical knowledge and ensures a reproducible and principled partitioning process. All documents within the databases are converted into high-dimensional vectors through a pre-trained and frozen multimodal encoder \(E_{\mathrm{retrieval}}\).

\subsubsection{Multimodal query embedding and parallel retrieval}
Given a query pair \((I, Q)\), we first generate a joint embedding that integrates the visual and textual semantics. Specifically, we employ a pre-trained multimodal encoder \(E_{\mathrm{retrieval}}\) based on a querying transformer architecture. This encoder conditions on the question \(Q\) to extract the most relevant visual features from the image \(I\)~\cite{ke2024developmenttestingretrievalaugmented}, producing a unified multimodal query embedding \(q_{\mathrm{multi}} \in \mathbb{R}^p\), which is given by
\begin{equation}
    q_{\mathrm{multi}} = E_{\mathrm{retrieval}}(I, Q).
\end{equation}
Subsequently, we use \(q_{\mathrm{multi}}\) to perform a parallel approximate $k$-nearest neighbor search across all \(D\) domain-specific vector databases \(\{\mathcal{DB}_d\}\). For each domain \(d\), this process returns the top-\(K\) most relevant documents and their corresponding cosine similarity scores, denoted as \(\mathcal{N}_d = \{(\mathrm{doc}_{d,k}, \mathrm{sim}_{d,k})\}_{k=1}^K\).

\subsubsection{Score aggregation, normalization, and probabilistic expert selection}
After obtaining the top-\(K\) similarity scores from the parallel retrieval process, we implement a three-stage procedure, which is designed to ensure that the routing decision is robust, comparable across domains, and adaptively responsive to query relevance.

\begin{itemize}
    \item \textit{Stage 1. Score aggregation:} To fuse the top-\(K\) similarity scores into a single metric, which represents the overall relevance of a query to a domain, we employ the arithmetic mean owing to its simplicity, interpretability, and parameter-free nature~\cite{kuo-etal-2025-mmlf}. By incorporating evidence from all top-\(K\) retrieved documents, this approach provides a holistic and robust estimate of domain relevance.
    \item \textit{Stage 2. Cross-domain score normalization:} Directly comparing the raw aggregated score \(s_d\) across domains can introduce systematic bias, as each domain knowledge base may differ in scale, density, or content homogeneity. To ensure fair comparison across domains, we adopt $z$-score~\cite{ke2024developmenttestingretrievalaugmented} normalization to transform each score into a standardized value, which is given by
    \begin{equation}
        \tilde{s}_d = \frac{s_d - \mu_d}{\sigma_d + \epsilon},
    \end{equation}
    where the domain-specific statistics \(\mu_d\) and \(\sigma_d\) are pre-estimated on a held-out validation set~\cite{ke2024developmenttestingretrievalaugmented}, and $\epsilon$ is a pre-defined hyperparameter~\cite{ke2024developmenttestingretrievalaugmented}. This step aims to mitigate score variations attributable to database characteristics rather than query relevance.
    \item \textit{Stage 3. Probabilistic selection and complex query handling:} The normalized scores \( \tilde{s}_d \) are fed into a softmax function with temperature \(\tau\) to generate the final routing probability distribution \(P(d|I, Q)\)~\cite{Lee2024RouterRetrieverRO, zhou-etal-2024-unveiling}, which is expressed as
    \begin{equation}
        P(d | I, Q) = \frac{\exp(\tilde{s}_d / \tau)}{\sum_{d'=1}^D \exp(\tilde{s}_{d'} / \tau)}.
    \end{equation}
    When the routing probability distribution indicates high uncertainty, the inference paradigm can be configured to activate multiple experts simultaneously, allowing collaborative inference and reducing the risk of suboptimal expert selection.

\end{itemize}

\subsection{Federated Governance for Adaptive Reasoning}
\label{ssubsec:fv_acot}

After selecting a context-augmented expert \(E_{d^*}\), the remaining challenge lies in determining when to terminate its iterative CoT reasoning. To this end, we introduce a federated governance mechanism that evaluates the confidence of the expert not only in its final prediction but also in the consistency and stability of its intermediate reasoning process.

\subsubsection{Meta-cognitive confidence estimation} The core of our governance mechanism is the meta-cognitive confidence estimator \(g_\psi: \mathbb{R}^h \to [0, 1]\), parameterized by \(\psi\)~\cite{gu2025harmonious}. This estimator provides a robust confidence score that reflects the stability of the reasoning state of LVLMs, moving beyond simple predictive probabilities that are often poorly calibrated~\cite{murugesan2024robust}.

We first model the interdependencies among experts through an expert dependency graph \(\mathcal{G} = (\mathcal{V}, \mathcal{E})\)~\cite{nguyen2025rem}, where the nodes \(\mathcal{V}=\{1, \ldots, D\}\) represent experts, and each directed edge \((i, j) \in \mathcal{E}\) signifies that expert \(i\) exerts a strong positive influence on the performance of expert \(j\). This interdependency is quantified by an influence score, expressed as~\cite{nguyen2025rem}
\begin{equation}
\mathrm{IS}(i \to j) = \mathrm{Acc}(j | \mathcal{M}_i) - \mathrm{Acc}(j | \mathcal{M}_{\mathrm{base}}),
\label{eq:influence_score}
\end{equation}
where \(\mathrm{Acc}(j | \mathcal{M}_i)\) represents the accuracy of expert \(j\) when its model \(\mathcal{M}\) is augmented with knowledge transferred from expert \(i\), and \(\mathrm{Acc}(j | \mathcal{M}_{\mathrm{base}})\) is its baseline accuracy.

The estimator \(g_\psi\) is implemented as a Multilayer Perceptron (MLP) consisting of a single hidden layer followed by a sigmoid activation function. The expert dependency graph \(\mathcal{G}\) is constructed by estimating the influence score for each expert pair \((i, j)\) on a held-out validation set~\cite{NEURIPS2024_7a8e7fd2}. To ensure the robustness of the estimated influence scores, we compute the average influence gain over \(K\) random restarts, as given by
\begin{equation}
\widehat{\mathrm{IS}}(i \to j) = \frac{1}{K} \sum_{r=1}^{K} \left( \mathrm{Acc}^{(r)}(j | \mathcal{M}_i) - \mathrm{Acc}^{(r)}(j | \mathcal{M}_{\mathrm{base}}) \right).
\end{equation}

A directed edge \((i, j)\) is added to \(\mathcal{G}\) only when the mean gain is statistically significant and exceeds a predefined threshold \(\tau\). Any cycle is resolved by greedily pruning the edge with the lowest gain. The estimator \(g_\psi\) takes the hidden state of the decoder \(h_t\) as input, and computes the final confidence score through two sequential stages:
\begin{itemize}
    \item \textit{Stage 1. Base confidence prediction:} The MLP sub-module \(g_{\mathrm{base}}\) first computes an initial confidence score directly from the hidden state $h_t$, i.e., \(u_{\mathrm{base}} = g_{\mathrm{base}}(h_t)\).
    
    \item \textit{Stage 2. Reasoning stability adjustment:} This base score $u_{\mathrm{base}}$ is then adjusted based on the reasoning stability~\cite{chen-etal-2023-ptp}. We identify the most influential parent from \(\mathcal{G}\), which is given by
    \begin{equation}\label{influential_parent}
        k = \argmax_{i \in \mathrm{Pa}(d^*)} \widehat{\mathrm{IS}}(i \to d^*),
    \end{equation}  
    where the mean feature representation \(\mu_k\) is either maintained locally at each site or obtained via secure aggregation. We then create a perturbed hidden state \(h'_t\) via interpolation, which is expressed as
    \begin{equation}
        h'_t = (1 - \epsilon)h_t + \epsilon \mu_k.
        \label{eq:perturbation_operator_impl}
    \end{equation}
    The stability adjustment is the negative and normalized Jensen-Shannon divergence between the output distributions conditioned on the original and perturbed states:
    \begin{equation}
        \Delta_{\mathrm{stability}} = - \frac{\mathrm{JS}\left( P(Y | h_t) \parallel P(Y | h'_t) \right)}{s},
        \label{eq:stability_adjustment_impl}
    \end{equation}
    where \(s\) is a normalization constant and $Y$ denotes the output variable of the LVLM.
\end{itemize}
The final stability-adjusted confidence is then given by \( c_{i,t} = \sigma\left(u_{\mathrm{base}} + \alpha \Delta_{\mathrm{stability}}\right) \), where $\alpha$ is a weight coefficient. To directly supervise the estimator, we introduce a regularization term $\mathcal{L}_{\mathrm{uncertainty\_reg}}$ that encourages alignment between its uncertainty predictions and the actual performance of the LVLM. Specifically, we define a target uncertainty \(u^*_t\) based on the correctness of the prediction at step \(t\):
\begin{equation}
    u^*_t = 
    \begin{cases} 
        u_{\mathrm{low}}, & \text{if } \argmax P(Y|h_t) = y_t, \\
        u_{\mathrm{high}}, & \text{otherwise},
    \end{cases}
\end{equation}
where \(u_{\mathrm{low}}\) and \(u_{\mathrm{high}}\) are hyperparameters representing low and high uncertainty, respectively. Hence, the regularization term is defined as a mean squared error loss that penalizes biases for the underlying uncertainty of estimators \(1 - u_{\mathrm{base}}\) from this target, which is given by
\begin{equation}
    \mathcal{L}_{\mathrm{uncertainty\_reg}} = \left( (1 - g_{\mathrm{base}}(h_t)) - u^*_t \right)^2.
\end{equation}
This loss directly trains the estimator \(g_{\mathrm{base}}\) to assign higher uncertainty to hidden states that are more likely to produce incorrect predictions.

%
\begin{algorithm}[t]
\small
\caption{LVLM Inference Paradigm in MedAlign}
\label{alg:medalign_inference_final_revised}
\begin{algorithmic}[1] 

\State \textbf{Input:} Query pair \((I, Q)\), set of experts \(\{E_d\}_{d=1}^D\), set of \(N\) federated sites, quorum size \(M \le N\), halting threshold \(\theta\), and maximum reasoning length \(T_{\max}\).
\State \textbf{Output:} Final generated answers \(A^*\).

\State \textcolor{blue}{\textit{\#\#\# Phase 1: Retrieval-aware expert routing \#\#\#}}
\State \textbf{Initialize:} Pre-computed expert dependency graph \(\mathcal{G}\) and meta-cognitive confidence estimator \(g_\psi\).

\State Retrieve context \(\{\mathcal{R}_d\}_{d=1}^D\) for each domain \(d\) from its knowledge base \(\mathcal{DB}_d\).
\State Compute normalized gating probabilities \(\{P(d|I, Q)\}_{d=1}^D\).
\State Select the optimal expert \(E_{d^*}\) where \(d^* = \argmax_d P(d|I, Q)\).

\State \textcolor{blue}{\textit{\#\#\# Phase 2: \textcolor{blue}Federated governance for adaptive reasoning} \#\#\#}
\State \textbf{Initialize:} Augmented context \(\mathrm{ctx}_0 \gets \mathrm{prompt}(I, Q, \mathcal{R}_{d^*})\).

\For{\(i = 1 \text{ to } N\)}
    \State \(\mathrm{ctx}_0^{(i)} \gets \mathrm{ctx}_0\).
\EndFor

\For{\(t = 1 \text{ to } T_{\max}\)}
    \State Get internal hidden state \(h_t\) from \(E_{d^*}\) given \(\mathrm{ctx}_{t-1}^{(i)}\).
    \State Compute base confidence \(u_{\mathrm{base}} \gets g_{\mathrm{base}}(h_t)\).
    \State Identify the most influential parent using Eq. (\ref{influential_parent}).
    \State Create a perturbed state using Eq.~\eqref{eq:perturbation_operator_impl}.
    \State Calculate adjustment using Eq.~\eqref{eq:stability_adjustment_impl}.
    \State Obtain the final meta-cognitive confidence.
    \If{\(c_{i,t} \ge \theta\)}
        \State Extract final answer \(A_i\) from reasoning chain \(\mathrm{ctx}_t^{(i)}\).
        \State Send tuple \((\mathrm{ctx}_t^{(i)}, A_i, c_{i,t})\) to the central server.
        \State \textbf{break}
    \EndIf
    \State Generate the next token \(y_t \sim E_{d^*}(\cdot | \mathrm{ctx}_{t-1}^{(i)})\).
    \State Append the token to context: \(\mathrm{ctx}_t^{(i)} \gets \mathrm{ctx}_{t-1}^{(i)} + y_t\).
\EndFor

\Statex \textcolor{blue}{\textit{\#\#\# Phase 3: Structured aggregation \#\#\#}}

\State \textbf{Initialize:} The set of confident reasoning chains $\mathcal{C}$.

\State Denote the consensus cluster $\mathcal{S}$ as \(\{A_i \text{ for } (\_, A_i, \_) \in \mathcal{C}\}\).
\If{a supermajority cluster exists in \(\mathcal{S}\)}
    \State Get the highest confidence answer $A^*$.
\Else
    \State Create the synthesis prompt $\{ \mathrm{CoT}_i \text{ for } (\mathrm{CoT}_i, \_, \_) \in \mathcal{C} \}$.
    
    \State Generate the final answer \(A^* \gets E_{d^*}(\cdot | \mathrm{prompt}_{\mathrm{synth}})\).
\EndIf

\State \Return \(A^*\)

\end{algorithmic}
\end{algorithm}


\subsubsection{Federated halting and structured aggregation}
In each reasoning step \(t\), each site computes its confidence score \(c_{i,t}\) and determines its halting decision \(d_{i,t}\). The global reasoning process terminates at step \(t^*\) once a quorum of \(M\) sites have halted. The final and critical stage involves aggregating the outputs from these \(M\) high-confidence reasoning chains. To ensure consistency and resolve potential conflicts among them, we design a structured two-stage aggregation procedure.
\begin{itemize}
    \item \textit{Stage 1. Consensus clustering and vetting:} The final answers \(\{A_1, A_2, \ldots, A_M\}\) from the halted sites are first embedded into a shared semantic space using a sentence encoder. We then perform density-based clustering to identify consensus groups. Specifically, if a single cluster contains a supermajority of the sites, its consensus is adopted~\cite{carrillo2024fedcbo}, and the final answer is selected from the reasoning chain of the site with the highest confidence score \(c_{i,t^*}\) within that cluster; otherwise, there exists a substantial reasoning conflict among the experts.
    \item \textit{Stage 2. Conflict resolution via synthesis:} In cases of conflict, we enter a synthesis phase. The complete reasoning chains \(\{\mathrm{CoT}_1, \mathrm{CoT}_2, \ldots, \mathrm{CoT}_M\}\) are concatenated into a single context block, which is presented by a meta-prompt. This prompt instructs an independent LLM instance to act as an impartial reviewer. The prompt is structured~\cite{zheng2025fedcalm} as ``\textit{Given the following conflicting reasoning paths from multiple experts attempting to answer the question `\{Q\}', analyze the logic, identify potential errors or hallucinations in each path, and synthesize a final, conclusive answer based on the most plausible evidence.}''
\end{itemize}
This aggregation procedure ensures that straightforward consensus cases are resolved efficiently, while complex disagreements are addressed through a transparent and explainable meta-reasoning process, rather than a naive voting mechanism. 

Algorithm~\ref{alg:medalign_inference_final_revised} illustrates the pseudocode of the LVLM inference paradigm in MedAlign. Its computational complexity is
\(\mathcal{O}(D(R + G) + N\bar T(\mathcal F(S)+\bar\Delta) + |\mathcal C|\log |\mathcal C| + \mathcal F(S_{\mathrm{synth}}) + |\mathcal C|S)\), where $R$ represents the average cost of retrieving content from \(\mathcal{DB}_d\), $G$ represents the average cost of computing a single domain gating \(P(d|I,Q)\), $\bar T$ represents the average number of halting steps per site, $\mathcal F(S)$ represents the model forward cost per step at context length \(S\), $S$ represents the average context length, $\bar\Delta$ represents the average number of parent nodes checked in the dependency graph \(\mathcal G\), and $S_{\mathrm{synth}}$ represents the average length of the synthesis prompt.

\section{Simulation Results}
\label{sec:experiments}

We conduct comprehensive experiments to evaluate the superiority of MedAlign. These experiments are designed to systematically answer five central research questions:
\begin{itemize}
    \item \textit{(Q1) SOTA Performance:} Does MedAlign outperform existing SOTA and strong baseline methods on diverse Med-VQA benchmarks under different learning settings?
    \item \textit{(Q2) Component Efficiency:} How do the core components of MedAlign, i.e., mDPO for grounding, RA-MoE for expert selection, and federated governance for adaptive reasoning, contribute to the overall performance?
    \item \textit{(Q3) Retrieval Effectiveness:} How does the proposed retrieval-aware gating compare with standard sparse, dense, and hybrid retrieval methods?
    \item \textit{(Q4) Robustness:} How robust is MedAlign to real-world data imperfections, such as noise or inconsistencies in visual and textual inputs?
    \item \textit{(Q5) Federated Governance Mechanism Validation:} Does the meta-cognitive uncertainty estimator effectively identify challenging or ambiguous cases, and how does the federated governance mechanism influence reasoning efficiency and overall accuracy?
\end{itemize}

\subsection{Experimental Setup}

\subsubsection{Experimental datasets} We evaluate MedAlign on three widely used Med-VQA benchmarks: MIMIC-CXR~\cite{Johnson2019}, IU-Xray~\cite{DemnerFushman2016}, and Harvard-FairVLMed~\cite{luo2024fairclip}. These datasets collectively cover diverse medical imaging modalities and question types, providing a comprehensive evaluation. Each dataset is partitioned into $80\%/10\%/10\%$ for training, validation, and testing, respectively.

\subsubsection{Evaluation metrics} For VQA tasks, we report accuracy and F1-score. To evaluate text generation quality, we employ BLEU~\cite{10.3115/1073083.1073135}, ROUGE-L~\cite{lin-2004-rouge}, and METEOR~\cite{banerjee-lavie-2005-meteor}. Efficiency is measured using the average inference latency.

\begin{table}[t]
\renewcommand{\arraystretch}{1.1}
\centering
\caption{Key implementation details of all experiments.}
\label{tab:hyperparams}
\begin{tabular}{ll}
\toprule[1.3pt]
\rowcolor{gray!10}
\textbf{Parameters} & \textbf{Values} \\
\midrule
\rowcolor{blue!10} 
\multicolumn{2}{l}{\textbf{Training Configuration}} \\
Optimizer & AdamW \\
Learning rate & \(1 \times 10^{-4}\) \\
Learning rate scheduler & Linear with $10\%$ warm-up \\
Weight decay & $0.01$ \\
Batch size & $16$ \\
Training epochs & $50$ \\
\midrule
\rowcolor{blue!10} 
\multicolumn{2}{l}{\textbf{Model Architecture}} \\
Base model & LLaVA-Med-1.5 7B \\
Vision encoder & CLIP ViT-L/14 (Frozen) \\
Adapter type & LoRA \\
LoRA rank (\(r\)) & $16$ \\
LoRA alpha (\(\alpha\)) & $32$ \\
\midrule
\rowcolor{blue!10} 
\multicolumn{2}{l}{\textbf{Framework-Specific Setup}} \\
Inverse temperature parameter (\(\beta\)) & $1.0$ \\
RA-MoE retrieved images (\(k\)) & $5$ \\
RA-MoE retrieved texts (\(m\)) & $5$ \\
Federated clients (\(N\)) & $5$ \\
Federated quorum (\(M\)) & $3\:(\lfloor N/2 \rfloor + 1)$ \\

\bottomrule[1.3pt]
\end{tabular}
\end{table}

\begin{table*}[t]
\renewcommand{\arraystretch}{1.2}
\centering
\caption{Comprehensive performance evaluation of MedAlign against all baselines on three Med-VQA datasets. We report standard VQA metrics (i.e., accuracy, precision, recall, and F1-score), text generation quality (i.e., BLEU, ROUGE-L, and METEOR), and inference efficiency (i.e., latency). Best results are highlighted in \textbf{bold}.}
\label{tab:grouped_metrics}

\resizebox{\textwidth}{!}{
\begin{tabular}{
    >{\centering\arraybackslash}m{2.5cm}|   
    >{\centering\arraybackslash}m{2.8cm}|   
    *{8}{>{\centering\arraybackslash}m{1.4cm}} 
}
\toprule[1.3pt]
\multirow{2}{*}{\textbf{Datasets}} & \multirow{2}{*}{\textbf{Methods}} 
& \multicolumn{4}{c}{\textbf{VQA Metrics}} 
& \multicolumn{3}{c}{\textbf{Generation Metrics}} 
& \multicolumn{1}{c}{\textbf{Efficiency}} \\
\cmidrule(lr){3-6} \cmidrule(lr){7-9} \cmidrule(lr){10-10}
\rowcolor{gray!10}
& & Acc (\%) & Rec & F1 (\%) & Prec (\%) & BLEU (\%) & ROUGE-L (\%) & METEOR (\%) & Latency ($\rm{s}$) \\
\midrule
\multirow{6}{*}{Iu-Xray} 
& LLaVA-Med           & 79.07  & 54.68  & 54.13  & 53.87  & 8.53  & 10.54  & 7.11  & 1.42  \\
\cmidrule(lr){2-10}
& + CoT     & 86.29  & 79.74  & 80.82  & 82.15  & 16.70  & 22.73  & 11.01   & 2.26  \\
& + BM25-RAG      & 70.99  & 77.16  & 73.62  & 70.92  & 11.31  & 13.93  & 9.32  & \textbf{1.24}  \\
& + Dense-RAG     & 77.47  & 78.55  & 75.18  & 72.99  & 15.51  & 17.88  &  11.26 &  1.32 \\
& + Hybrid-RAG    & 86.58  & 85.96  & 83.16  & 81.41  & 18.60  & 22.28  & 13.80  & 1.87  \\
\cmidrule(lr){2-10}
\rowcolor{blue!10}
& \textbf{MedAlign} & \textbf{96.30}  & \textbf{93.58}  & \textbf{95.01}  & \textbf{96.70}  & \textbf{22.41}  & \textbf{28.06}  & \textbf{18.36}  & 2.14  \\
\midrule
\multirow{6}{*}{Harvard-FairVLMed} 
& LLaVA-Med           & 55.03  & 65.00  & 61.93  & 59.13  & 11.66  & 14.07  & 9.66  &  \textbf{1.37} \\
\cmidrule(lr){2-10}
& + CoT     & 83.54  & 65.33  & 64.81  & 64.35  & 17.44  & 21.67  & 15.12  & 4.88 \\
& + BM25-RAG      & 83.41  & 69.28  & 67.80  & 66.70  & 13.37  & 16.94  & 12.40  & 2.65  \\
& + Dense-RAG     & 62.59  & 69.58  & 65.41  & 61.11  & 14.38  & 17.54  &  12.23 & 1.45  \\
& + Hybrid-RAG    & 78.05  & 72.36  & 67.63  & 63.62  & 17.67  & 22.20  & 15.34  & 2.41  \\
\cmidrule(lr){2-10}
\rowcolor{blue!10}
& \textbf{MedAlign} &  \textbf{97.73} & \textbf{96.73}  & \textbf{94.96}  & \textbf{93.36}  & \textbf{23.40}  & \textbf{29.30}  & \textbf{16.79}  &  2.62 \\
\midrule
\multirow{6}{*}{MIMIC-CXR} 
& LLaVA-Med           & 67.76  & 68.89  & 68.03  & 67.88  & 11.50  & 13.43  &  8.58 & 1.68  \\
\cmidrule(lr){2-10}
& + CoT     & 71.88  & 70.35  & 71.56  & 72.11  & 15.63  & 18.09  & 11.63  & 4.66  \\
& + BM25-RAG      & 75.77  & 70.19  & 73.26  & 77.24  & 12.97  & 15.38  & 12.50  & \textbf{1.43}  \\
& + Dense-RAG     & 70.35  & 69.29  & 70.19  & 71.50  & 13.95  & 18.23  & 11.51  & 1.75  \\
& + Hybrid-RAG    & 78.12  & 76.71  & 76.68  & 76.66  & 15.91  & 18.90  & 12.82  & 3.17  \\
\cmidrule(lr){2-10}
\rowcolor{blue!10}
& \textbf{MedAlign} & \textbf{95.64}  & \textbf{96.47}  & \textbf{95.25}  & \textbf{94.36}  & \textbf{22.99}  & \textbf{27.89}  & \textbf{16.33}  & 2.29  \\
\bottomrule[1.3pt]
\end{tabular}
}
\end{table*}

\subsection{Performance Comparison Setting}

To rigorously evaluate the performance of MedAlign, we compare it with a comprehensive suite of baselines, including foundational LVLMs, previous SOTA retrieval paradigms, and ablation variants. We introduce them as follows:

\begin{itemize}
    \item \textbf{Foundational Models:} Our primary baseline is the foundational LLaVA-Med-1.5 7B, fine-tuned directly on the clinical datasets, which enables precise evaluation of the performance improvements introduced by mDPO.

    \item \textbf{Retrieval-Augmented Baselines:} To validate the superiority of our RA-MoE architecture, we implement several strong baselines built upon the RAG paradigm~\cite{10.5555/3495724.3496517}.
        \begin{itemize}
            \item \textit{BM25-RAG:} Utilize a sparse retriever based on the classic BM25 probabilistic relevance model~\cite{10.1561/1500000019}.
            \item \textit{Dense-RAG:} Employ a dense retriever, which is implemented following the principles of dense passage retrieval~\cite{10.5555/3495724.3496517}. For multimodal context, our implementation is inspired by recent vision-centric RAG approaches~\cite{yu2025visrag}.
            \item \textit{Hybrid-RAG:} Combine sparse and dense retrieval scores. This baseline represents a strong contemporary approach, as recent studies have demonstrated that hybrid retrieval methods often outperform single-retriever systems~\cite{chen2024dense}. Our score fusion strategy follows recent advances in efficient hybrid vector retrieval~\cite{Zhang2024EfficientAE}.
        \end{itemize}

    \item \textbf{Sanity Verification Baseline:}
        \begin{itemize}
            \item \textit{CoT Only:} To separately measure the impact of reasoning-eliciting prompts, the basic LLaVA-Med model is prompted with a standard CoT instruction, following the seminal methodology established by~\cite{NEURIPS2022_9d560961}. The prompt structure is specifically adapted for the visual domain, guided by best practices from recent visual CoT benchmarks~\cite{shao2024visualcotadvancingmultimodal}.
        \end{itemize}

    \item \textbf{Ablation Variants of MedAlign:} To dissect the contribution of each component within MedAlign, we evaluate several ablation variants:
        \begin{itemize}
            \item \textit{MedAlign (w/o mDPO):} We fine-tune the LVLM without the mDPO objective to isolate the contribution of our fine-tuning strategy.
            \item \textit{MedAlign (w/o RA-MoE):} We leverage a simple text-based gating architecture instead of our visually-grounded routing.
            \item \textit{MedAlign (w/o Meta-cognitive adjustment):} We utilize the complete federated halting mechanism, replacing the full meta-cognitive uncertainty estimator with a simpler confidence score.
            \item \textit{MedAlign (Fixed-depth CoT):} We disable adaptive reasoning, where CoT reasoning proceeds to a fixed maximum length.
        \end{itemize}
\end{itemize}



\subsection{Implementation Details}
\label{subsec:implementation_details}

Our framework is built upon the LLaVA-Med-1.5 7B model. To ensure fair comparisons, all models and baselines are trained under a unified protocol, with key implementation details summarized in Table~\ref{tab:hyperparams}. Specifically,

\begin{table}[t]
\renewcommand{\arraystretch}{1.2}
\centering
\caption{Ablation study on MedAlign.}
\label{tab:ablation}

\begin{tabular}{lccc}
\toprule[1.3pt]
\rowcolor{gray!10}
\textbf{Model Variants} & {Acc (\%)} & {F1 (\%)} & {t ($\rm{s}$)} \\
\midrule
\rowcolor{blue!10} 
\textbf{MedAlign} & \textbf{96.30}  & \textbf{95.01} & 2.14 \\
\cmidrule(lr){1-4}
w/o mDPO              & 88.08 & 84.62 & 4.37 \\
w/o RA-MoE             & 92.49 & 90.11 & \textbf{1.99} \\
w/o Meta-cognitive adjustment & 90.92 & 88.66 & 2.09 \\
\bottomrule[1.3pt]
\end{tabular}

\vspace{0.5 em} 
\subcaption{Iu-Xray}

\vspace{0.8 em} 

\begin{tabular}{lccc}
\toprule[1.3pt]
\rowcolor{gray!10}
\textbf{Model Variants} & {Acc (\%)} & {F1 (\%)} & {t ($\rm{s}$)} \\
\midrule
\rowcolor{blue!10} 
\textbf{MedAlign} & \textbf{97.73} & \textbf{94.96}  & 2.62 \\
\cmidrule(lr){1-4}
w/o mDPO              & 89.43 & 81.25 & 5.08 \\
w/o RA-MoE             & 94.36 & 87.76 & \textbf{2.28} \\
w/o Meta-cognitive adjustment & 91.43 & 84.67 & 2.44 \\
\bottomrule[1.3pt]
\end{tabular}
\vspace{0.5 em} 

\subcaption{Harvard-FairVLMed}

\vspace{0.8 em}

\begin{tabular}{lccc}
\toprule[1.3pt]
\rowcolor{gray!10}
\textbf{Model Variants} & {Acc (\%)} & {F1 (\%)} & {t ($\rm{s}$)} \\
\midrule
\rowcolor{blue!10} 
\textbf{MedAlign} & \textbf{95.64} & \textbf{95.25} & 2.29 \\
\cmidrule(lr){1-4}
w/o mDPO              & 86.13 & 85.52 & 3.30 \\
w/o RA-MoE             & 93.19 & 92.64 & \textbf{2.16}\\
w/o Meta-cognitive adjustment & 90.69 & 90.12 & 2.25 \\
\bottomrule[1.3pt]
\end{tabular}
\vspace{0.5 em}
\subcaption{MIMIC-CXR}
\end{table}

\begin{itemize}
    \item \textbf{Architectural Choices.} All custom modules, including the mDPO preference heads, domain-specific RA-MoE experts, and the meta-cognitive uncertainty estimator, are implemented as parameter-efficient LoRA adapters. This design significantly reduces the number of trainable parameters. For visual processing, we employ the powerful CLIP ViT-L/14 vision encoder, which remains frozen throughout all fine-tuning stages to preserve its strong pre-trained representations.
    \item \textbf{Data and Simulation Setup.} Preference data for mDPO are constructed by pairing each ground-truth answer (preferred) with a high-probability but incorrect response generated by the base SFT model (dispreferred). For the robustness experiments, we systematically evaluate model resilience by injecting both textual and visual noise at varying levels from $0\%$ to $30\%$. The federated setting for the federated governance mechanism is simulated on a single node with \(N=5\) virtual clients.
\end{itemize}

\subsection{Overall Performance}
\label{subsec:overall_performance}

We first present the main results of MedAlign compared with all baselines on three representative Med-VQA benchmarks in Table~\ref{tab:grouped_metrics}. The results clearly demonstrate the superior performance of MedAlign across all evaluation metrics.

\textit{(A1) MedAlign achieves a new SOTA across all datasets.} On MIMIC-CXR, MedAlign attains an F1-score of $95.25\%$, surpassing the strongest baseline, Hybrid-RAG, by a significant margin of $18.57\%$. This trend holds for IU-Xray and Harvard FairVLMed, where MedAlign shows consistent improvements in all VQA metrics, i.e., accuracy, precision, recall, and F1-score. Furthermore, the high scores in BLEU, ROUGE-L, and METEOR indicate that the reasoning chains generated by MedAlign are not only factually correct but also linguistically fluent and coherent.

\textit{(A3) The designed retrieval-aware gating significantly outperforms standard RAG.} As shown in Table~\ref{tab:grouped_metrics}, MedAlign consistently outperforms all standard retrieval baselines, i.e., BM25-RAG, Dense-RAG, and Hybrid-RAG, which validates our core hypothesis that a visually-grounded expert-routing architecture is more effective than simply augmenting the prompt with retrieved text, as it allows the LVLM to leverage specialized knowledge. Notably, while MedAlign has a slightly higher inference latency than the non-retrieval baselines due to its components, it remains computationally competitive with the powerful Hybrid-RAG.

\subsection{Ablation Studies}
\label{subsec:ablation}

To deconstruct the sources of the performance and efficiency of MedAlign, we conduct a comprehensive ablation study across the Iu-Xray, Harvard-FairVLMed, and MIMIC-CXR datasets, with results presented in Table~\ref{tab:ablation}. This analysis provides strong evidence that each proposed component contributes significantly to the overall efficacy and robustness of the framework (for \textit{Q2}).

Even with the removal of any single module, MedAlign maintains a minimum accuracy of $86.13\%$ and an F1-score exceeding $81.25\%$, which demonstrates its robustness and highlights a strong synergistic effect among the architectural components. The most pronounced performance degradation is observed in the absence of our mDPO objective on the Harvard-FairVLMed dataset, which incurs a $13.71\%$ decline in the F1-score. This result highlights the critical role of our fine-tuning strategy in achieving robust visual grounding and representation stability. Similarly, replacing the visually-grounded RA-MoE with a standard gating architecture leads to a substantial erosion in performance, confirming the benefits of modality-aware expert selection.

The ablations on our adaptive reasoning process are particularly informative. The meta-cognitive uncertainty estimator serves as a critical component, contributing significantly to both accuracy and computational efficiency. By setting the confidence threshold to $0.8$, MedAlign achieves a compelling trade-off, matching the accuracy of the most computationally intensive fixed-depth CoT variant while substantially reducing inference time. This demonstrates that MedAlign can achieve SOTA accuracy with a significant advantage in efficiency, highlighting its practical value for clinical service deployment.

\begin{figure*}[htbp]
    \centering

    \begin{minipage}[b]{0.32\textwidth}
        \centering
        \includegraphics[width=\linewidth,
                         height=0.17\textheight,
                         keepaspectratio]{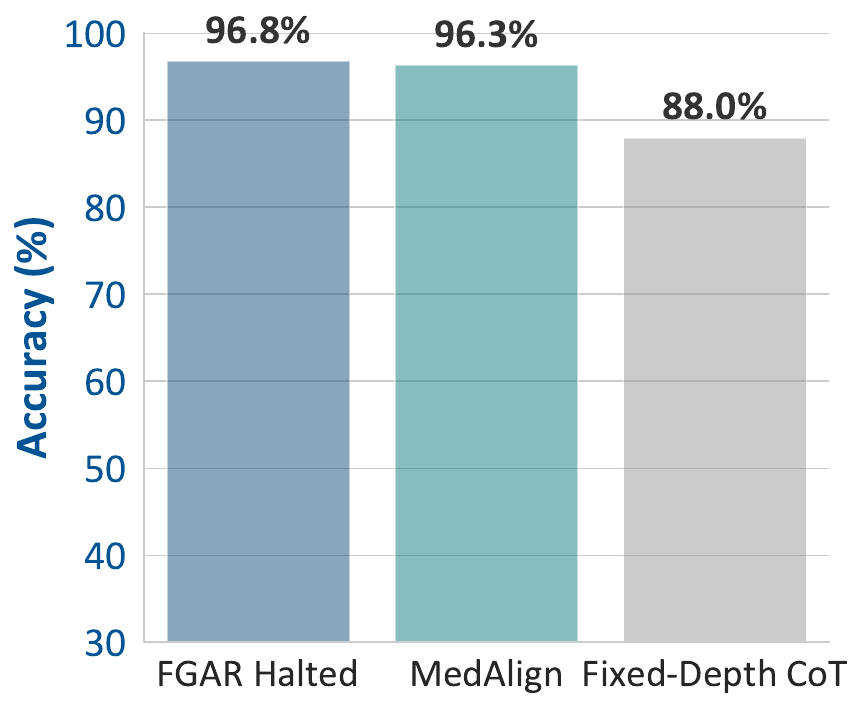}
        \subcaption{}\label{Iu_Xray_accuracy}
        \vspace{0.1cm}\par
        \includegraphics[width=\linewidth,
                         height=0.17\textheight,
                         keepaspectratio]{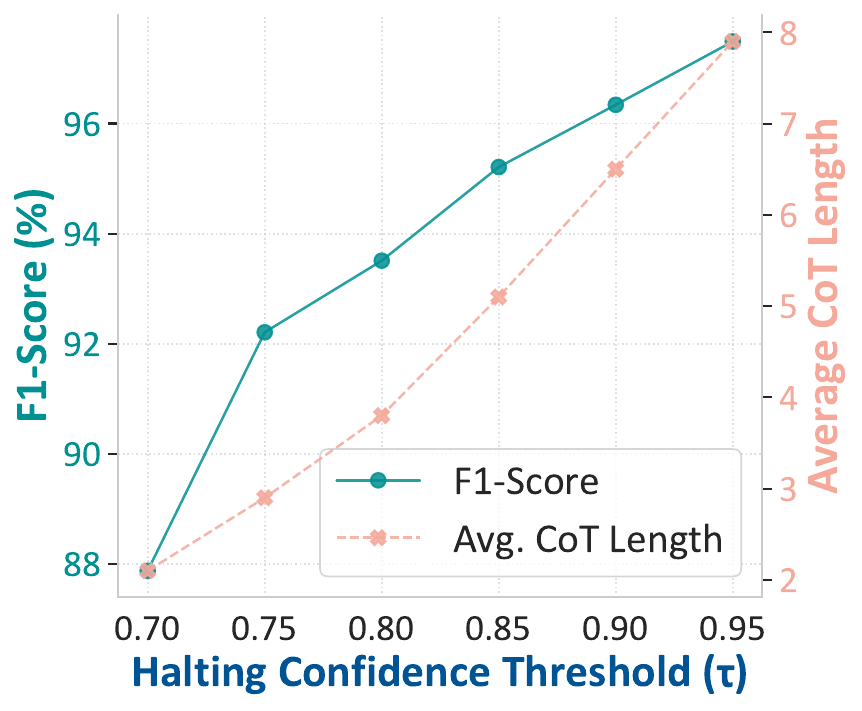}
        \subcaption{}\label{Iu_Xray_F1}
    \end{minipage}
    \hfill
    \begin{minipage}[b]{0.32\textwidth}
        \centering
        \includegraphics[width=\linewidth,
                         height=0.17\textheight,
                         keepaspectratio]{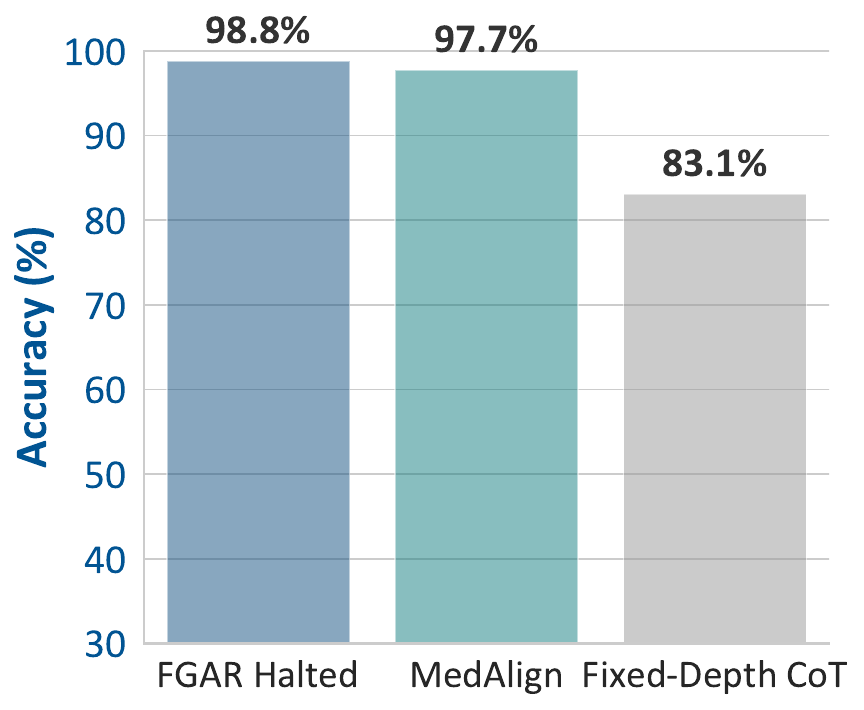}
        \subcaption{}\label{Harvard_FairVLMed_accuracy}
        \vspace{0.1cm}\par
        \includegraphics[width=\linewidth,
                         height=0.17\textheight,
                         keepaspectratio]{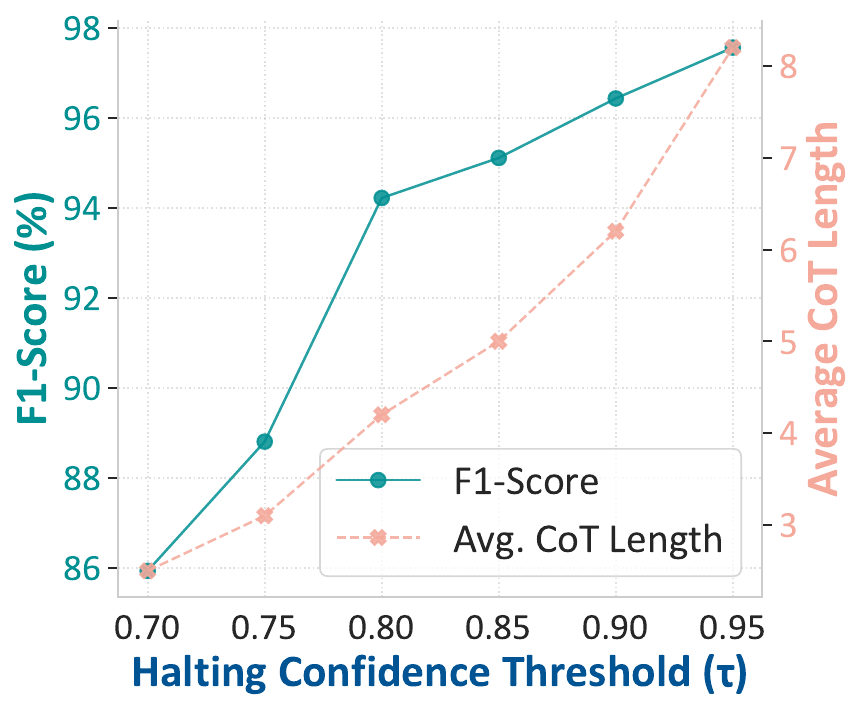}
        \subcaption{}\label{Harvard_FairVLMed_F1}
    \end{minipage}
    \hfill
    \begin{minipage}[b]{0.32\textwidth}
        \centering
        \includegraphics[width=\linewidth,
                         height=0.17\textheight,
                         keepaspectratio]{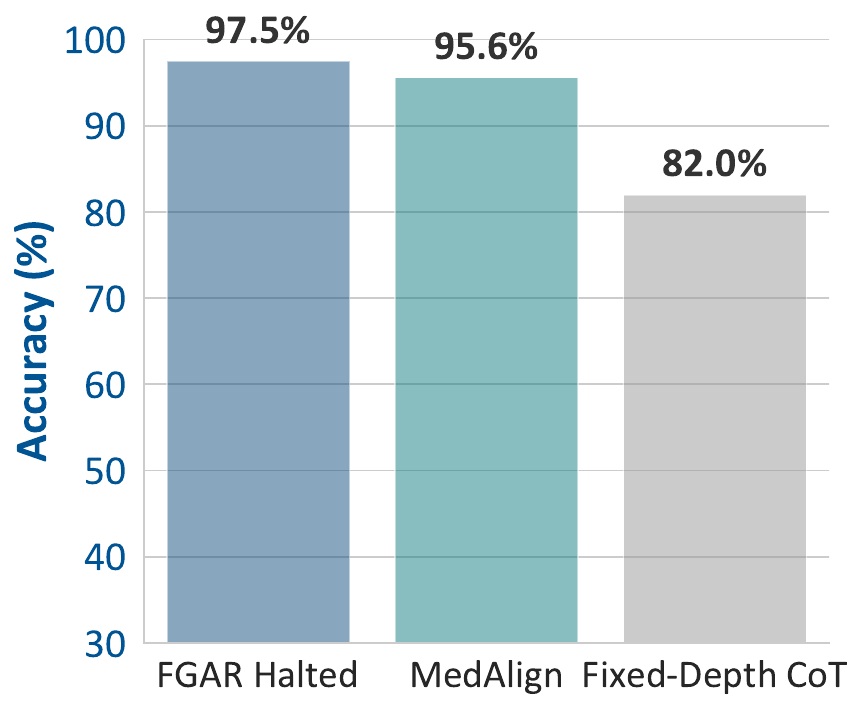}
        \subcaption{}\label{MIMIC_CXR_accuracy}
        \vspace{0.1cm}\par
        \includegraphics[width=\linewidth,
                         height=0.17\textheight,
                         keepaspectratio]{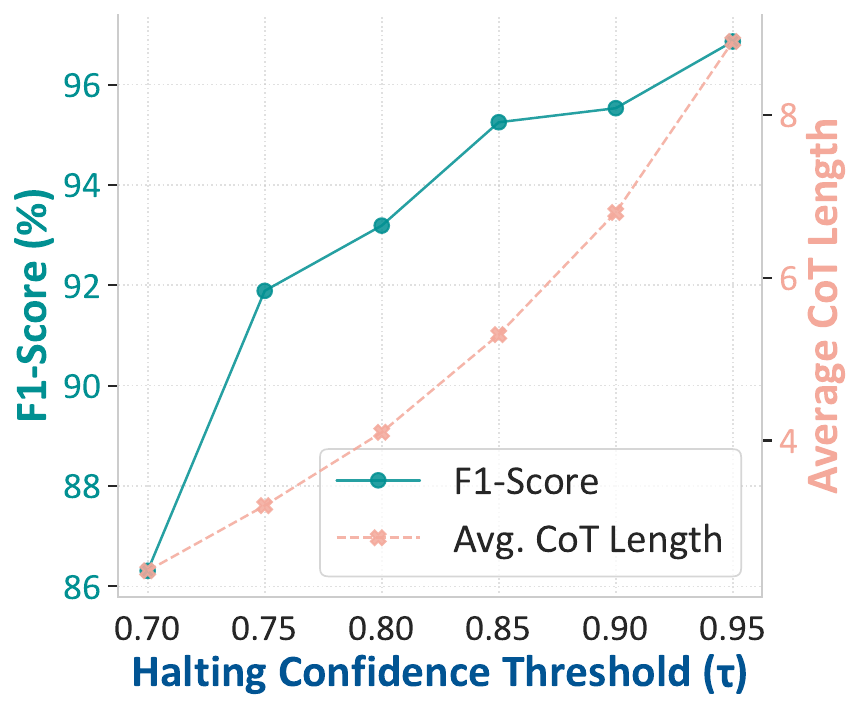}
        \subcaption{}\label{MIMIC_CXR_F1}
    \end{minipage}
    
    \makebox[0.32\textwidth][c]{\footnotesize Iu-Xray}\hfill
    \makebox[0.32\textwidth][c]{\footnotesize Harvard-FairVLMed}\hfill
    \makebox[0.32\textwidth][c]{\footnotesize MIMIC-CXR}

    \caption{
    Performance evaluation of the Federated Governance mechanism for Adaptive Reasoning (FGAR) across three clinical datasets: Iu-Xray, Harvard-FairVLMed, and MIMIC-CXR. Figures 2(a), 2(c), and 2(e) compare the accuracy of reasoning chains adaptively terminated by our halting criterion with those run to a fixed depth. The significant performance gap validates the effectiveness of our meta-cognitive uncertainty estimator. Figures 2(b), 2(d), and 2(f) illustrate the explicit trade-off between model performance (F1-score) and computational overhead (average CoT length) as a function of the confidence threshold \(\tau\).
}
    \label{fig:Metacognitive_analysis}
\end{figure*}

\begin{figure*}[htbp]
    \centering

    \begin{minipage}[b]{0.32\textwidth}
        \centering
        \includegraphics[width=\linewidth,
                         height=0.17\textheight,
                         keepaspectratio]{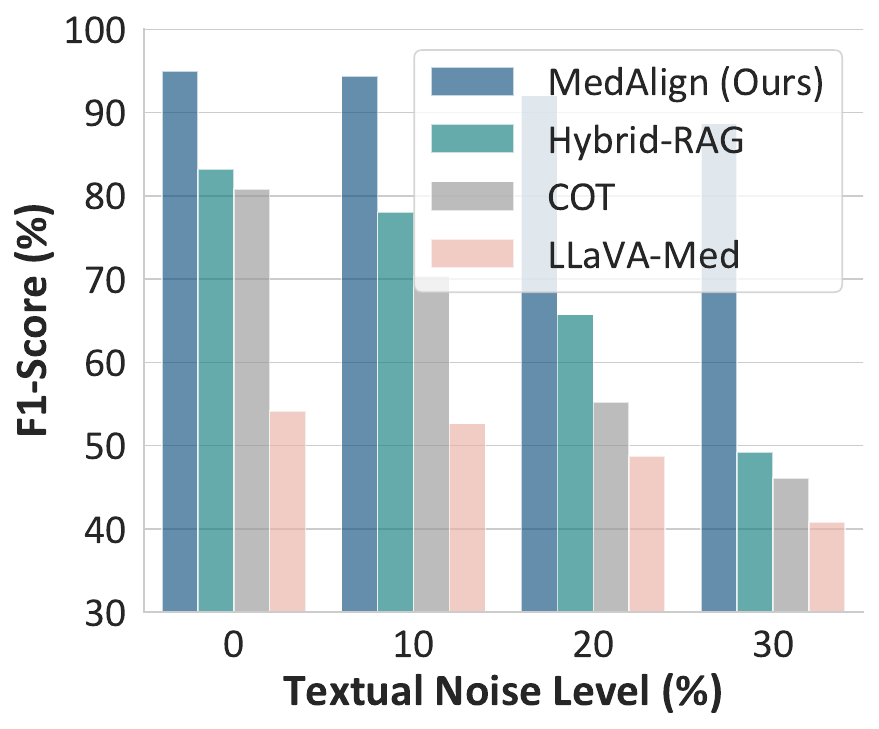}
        \subcaption{}\label{Iu_Xray_textual_noise}
        \vspace{0.1cm}\par
        \includegraphics[width=\linewidth,
                         height=0.17\textheight,
                         keepaspectratio]{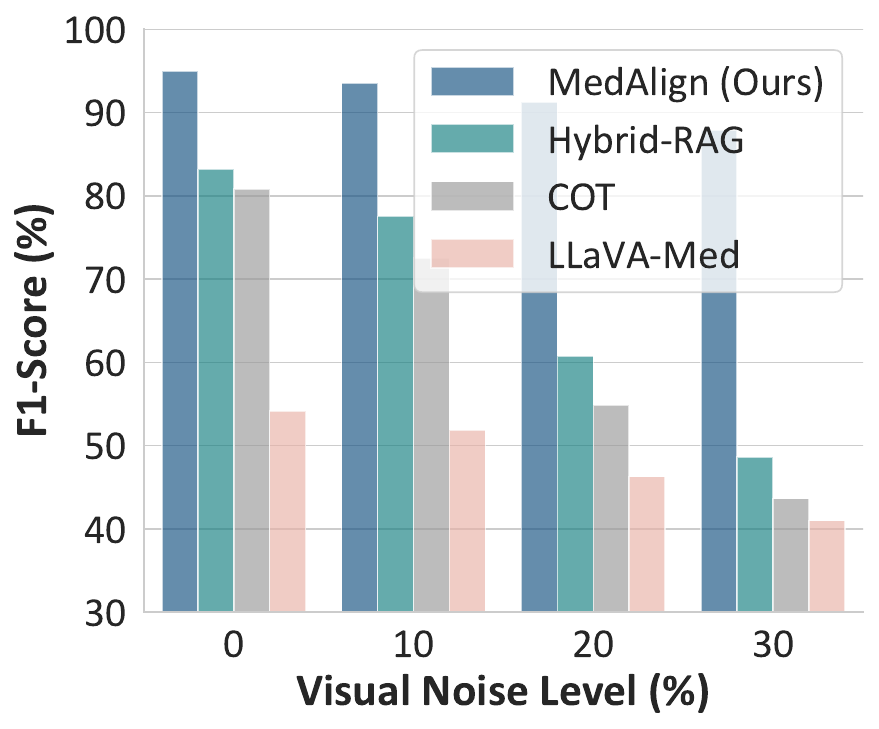}
        \subcaption{}\label{Iu_Xray_visual_noise}
    \end{minipage}
    \hfill
    \begin{minipage}[b]{0.32\textwidth}
        \centering
        \includegraphics[width=\linewidth,
                         height=0.17\textheight,
                         keepaspectratio]{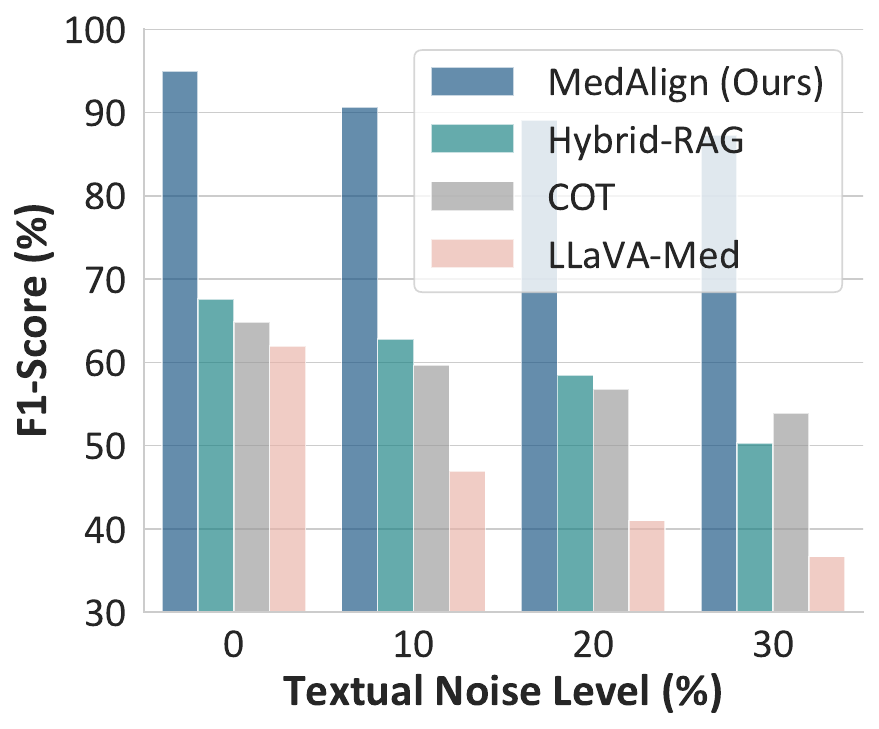}
        \subcaption{}\label{Harvard_FairVL_textual_noise}
        \vspace{0.1cm}\par
        \includegraphics[width=\linewidth,
                         height=0.17\textheight,
                         keepaspectratio]{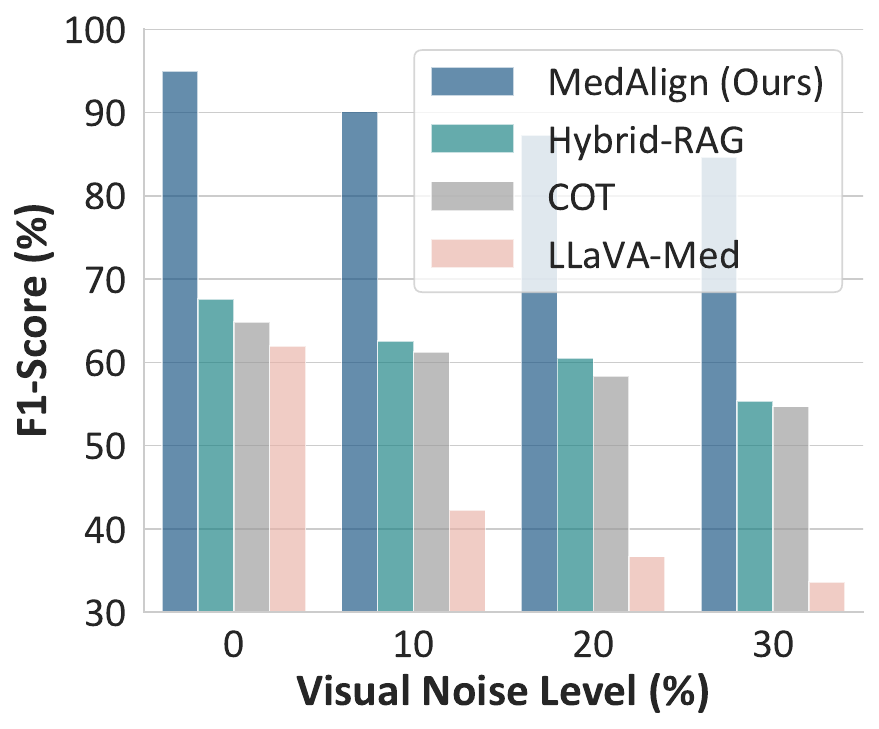}
        \subcaption{}\label{Harvard_FairVL_visual_noise}
    \end{minipage}
    \hfill
    \begin{minipage}[b]{0.32\textwidth}
        \centering
        \includegraphics[width=\linewidth,
                         height=0.17\textheight,
                         keepaspectratio]{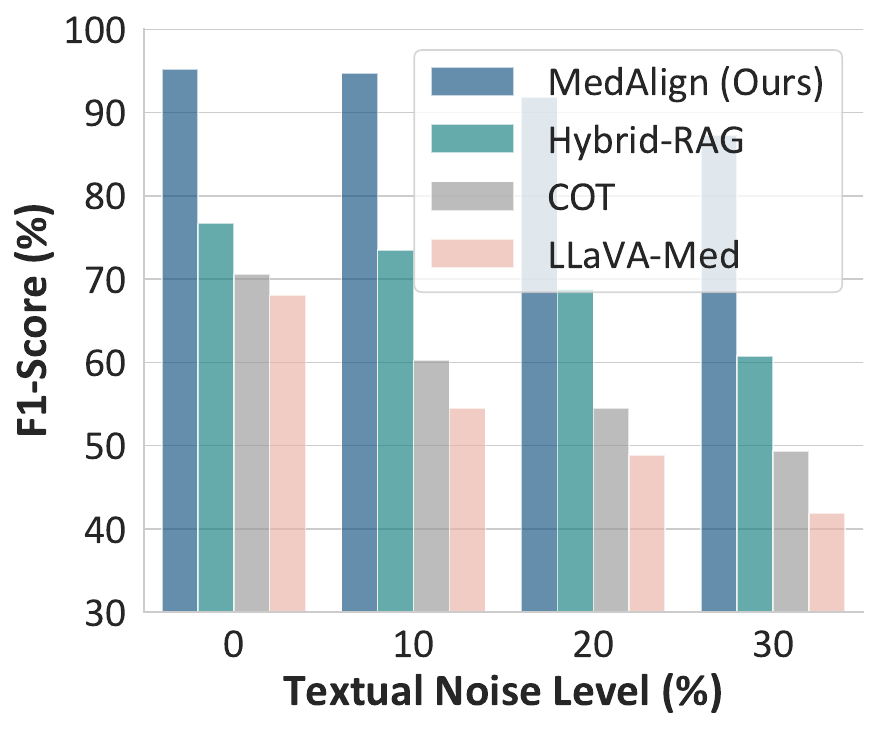}
        \subcaption{}\label{MIMIC_CXR_textual_noise}
        \vspace{0.1cm}\par
        \includegraphics[width=\linewidth,
                         height=0.17\textheight,
                         keepaspectratio]{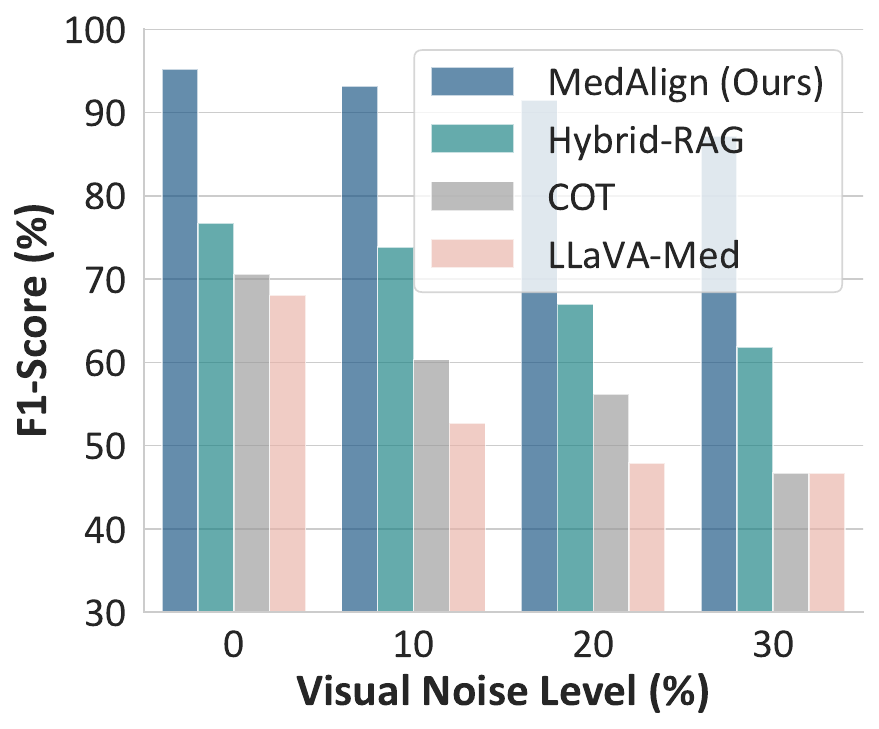}
        \subcaption{}\label{MIMIC_CXR_visual_noise}
    \end{minipage}
    
    \makebox[0.32\textwidth][c]{\footnotesize Iu-Xray}\hfill
    \makebox[0.32\textwidth][c]{\footnotesize Harvard-FairVLMed}\hfill
    \makebox[0.32\textwidth][c]{\footnotesize MIMIC-CXR}

    \caption{
    Robustness evaluation of MedAlign to data perturbations across three clinical datasets: Iu-Xray, Harvard-FairVLMed, and MIMIC-CXR. The F1-scores of MedAlign and baseline models are plotted against increasing noise intensity. Figures 3(a), 3(c), and 3(e) depict resilience to textual noise, simulated via token substitutions and deletions. Figures 3(b), 3(d), and 3(f) depict resilience to visual noise, simulated via Gaussian noise and random patch occlusion.
}
    \label{fig:robustness}
\end{figure*}

\subsection{Analysis of the Federated Governance Mechanism}
\label{subsec:Metacognitive_analysis}

To validate the effectiveness and controllability of our federated governance mechanism (for \textit{Q5}), we perform a detailed behavioral analysis, with results consistently replicated across all three datasets, as illustrated in Fig.~\ref{fig:Metacognitive_analysis}.

Figures~\ref{Iu_Xray_accuracy},~\ref{Harvard_FairVLMed_accuracy}, and~\ref{MIMIC_CXR_accuracy} provide strong evidence for the effectiveness of our meta-cognitive uncertainty estimator. Across all clinical datasets, reasoning chains adaptively terminated by our federated governance mechanism consistently achieve substantially higher accuracy than those extended to a fixed depth. For instance, on the MIMIC-CXR dataset, reasoning chains halted by the mechanism achieve an accuracy of $97.50\%$, compared with $82.00\%$ for chains that continued to the fixed depth. This large performance gap empirically demonstrates that our method effectively discerns high-confidence, correct reasoning paths from those that are uncertain or speculative. The performance of the fixed-depth CoT variant demonstrates the advantage of our dynamic reasoning mechanism over static approaches. Moreover, the large performance disparity between the full MedAlign and the variant w/o meta-cognitive adjustment provides direct evidence for the effectiveness of our meta-cognitive uncertainty formulation. Conventional confidence heuristics, such as maximum log-probability, are insufficient for reliable dynamic halting. In contrast, our federated governance mechanism plays a crucial role in robustly identifying well-grounded reasoning paths. Figures~\ref {Iu_Xray_F1}, \ref{Harvard_FairVLMed_F1}, and \ref{MIMIC_CXR_F1} illustrate the precise controllability of the federated governance mechanism. We observe that there exists a clear and monotonic relationship among the confidence threshold \(\tau\), the F1-score, and the average CoT length. Specifically, as \(\tau\) increases, indicating that higher certainty for early termination is required, the performance of the LLaVA-Med model systematically improves at the cost of increased computational steps. This tunable behavior demonstrates that the federated governance mechanism allows practitioners to specify the desired balance between diagnostic performance and inferential efficiency according to application-specific requirements.

\subsection{Robustness to Real-World Noise}
\label{subsec:robustness}

To evaluate the practical reliability of MedAlign, we conduct a systematic evaluation of its robustness against simulated real-world data imperfections (for \textit{Q4}). As depicted in Fig.~\ref{fig:robustness}, MedAlign exhibits substantially enhanced resilience to both textual and visual noise across all three benchmark datasets, outperforming all prominent baselines.

While the performance of all approaches predictably degrades with increasing noise intensity, the performance trajectory of MedAlign exhibits a notably shallower slope. Quantitatively, MedAlign retains over $85\%$ of its original F1-score even under $40\%$ noise levels. In stark contrast, the performance of the foundational LLaVA-Med model degrades precipitously under identical conditions, suggesting a reliance on fragile, surface-level data correlations. We attribute this heightened robustness to the synergistic interplay of MedAlign core components. The robust visual grounding introduced by the mDPO objective, combined with the context-aware expert selection of the RA-MoE architecture, enables MedAlign to possess a stronger ability to resist false interference in the input data. This result indicates that MedAlign is not only accurate under ideal conditions but also resilient and reliable in noisy, imperfect clinical environments, which is a crucial property for real-world clinical services.

\section{Conclusion}
\label{sec:conclusion}

In this paper, we have developed MedAlign, a novel framework to enable visually accurate LVLM responses for Med-VQA. Specifically, we have proposed mDPO to mitigate hallucinations in LVLMs, which explicitly aligns preference learning with visual context. Moreover, we have designed an RA-MoE architecture that dynamically activates the most contextually relevant LVLM expert, thereby addressing visual grounding failures. Furthermore, we have proposed a federated governance mechanism for adaptive reasoning, enabling each selected expert to autonomously regulate its CoT reasoning depth. This design effectively alleviates inefficient reasoning while facilitating efficient collaboration across multiple institutions. Extensive experiments demonstrate that MedAlign not only establishes new SOTA performance across representative Med-VQA benchmarks but also exhibits superior robustness against real-world data imperfections. For future work, we plan to extend MedAlign beyond VQA to a broader range of clinical tasks and enable collaborative inference among multiple experts, thereby advancing toward artificial intelligence systems that are not only powerful but also efficient and robust for real-world clinical services.

\bibliographystyle{IEEEtran}
\bibliography{references}
\end{document}